\documentclass[10pt,letterpaper,twocolumn]{article}

\usepackage{cvpr}
\usepackage{times}
\usepackage[accsupp]{axessibility}

\usepackage{color,xcolor}
\usepackage{epsfig}
\usepackage{graphicx}

\usepackage{adjustbox}
\usepackage{array}
\usepackage{booktabs}
\usepackage{colortbl}
\usepackage{float,wrapfig}
\usepackage{hhline}
\usepackage{multirow}
\usepackage{subcaption} % issues a warning with CVPR/ICCV format
\usepackage[toc,page]{appendix}

\usepackage{amsmath,amsfonts,amsthm,amssymb}
\usepackage{bm}
\usepackage{nicefrac}
\usepackage{microtype}
\usepackage{inconsolata}
\usepackage{pifont}

\usepackage{changepage}
\usepackage{extramarks}
\usepackage{fancyhdr}
\usepackage{lastpage}
\usepackage{setspace}
\usepackage{soul}
\usepackage{xspace}
\usepackage{indentfirst}
\usepackage{paralist}
\usepackage{booktabs,arydshln}

\usepackage[pagebackref=true,breaklinks=true,letterpaper=true,colorlinks,citecolor=citecolor,bookmarks=false]{hyperref}
\usepackage{url}

\usepackage{algorithm, algorithmic}
\usepackage{enumerate}
\usepackage{lipsum}
\newcolumntype{L}[1]{>{\raggedright\let\newline\\\arraybackslash\hspace{0pt}}m{#1}}
\newcolumntype{C}[1]{>{\centering\let\newline\\\arraybackslash\hspace{0pt}}m{#1}}
\newcolumntype{R}[1]{>{\raggedleft\let\newline\\\arraybackslash\hspace{0pt}}m{#1}}

\newcommand{\sect}[1]{Section~\ref{#1}}

\newcommand{\fig}[1]{Figure~\ref{#1}}

\newcommand{\tab}[1]{Table~\ref{#1}}

\newcommand{\ignorethis}[1]{}

\makeatletter
\DeclareRobustCommand\onedot{\futurelet\@let@token\@onedot}
\def\@onedot{\ifx\@let@token.\else.\null\fi\xspace}

\def\eg{\emph{e.g}\onedot} 
\def\ie{\emph{i.e}\onedot} 
 
 \def\vs{\emph{vs}\onedot}
\def\wrt{w.r.t\onedot} 

\makeatother

\makeatletter
\def\adl@drawiv#1#2#3{%
        \hskip.5\tabcolsep
        \xleaders#3{#2.5\@tempdimb #1{1}#2.5\@tempdimb}%
                #2\z@ plus1fil minus1fil\relax
        \hskip.5\tabcolsep}
\newcommand{\cdashlinelr}[1]{%
  \noalign{\vskip\aboverulesep
           \global\let\@dashdrawstore\adl@draw
           \global\let\adl@draw\adl@drawiv}
  \cdashline{#1}
  \noalign{\global\let\adl@draw\@dashdrawstore
           \vskip\belowrulesep}}
\makeatother

\definecolor{citecolor}{HTML}{0071bc}
\definecolor{mydarkblue}{rgb}{0,0.08,1}
\definecolor{mydarkgreen}{rgb}{0.02,0.6,0.02}
\definecolor{darkred}{rgb}{0.8,0.02,0.02}
\definecolor{darkorange}{rgb}{0.40,0.2,0.02}
\definecolor{darkpurple}{RGB}{111,0,255}
\definecolor{myred}{rgb}{1.0,0.0,0.0}
\definecolor{mygold}{rgb}{0.75,0.6,0.12}
\definecolor{mydarkgray}{rgb}{0.66, 0.66, 0.66}

\definecolor{spc}{RGB}{119, 107, 170}
\definecolor{pct}{rgb}{0.7, 0, 0.2}

\newcommand{\myparagraph}[1]{\vspace{-10pt}\paragraph{#1}}

\newcommand{\spc}{\textcolor{spc}{$\mathbf{\circ}$\,}}
\newcommand{\pct}{\textcolor{pct}{$\bullet$\,}}

\def\model{FlatFormer\xspace}

\def\cvprPaperID{9743}
\def\confName{CVPR}
\def\confYear{2023}

\begin{document}

\title{\model: \underline{Flat}tened Window Attention for Efficient Point Cloud Trans\underline{former}}

\author{
Zhijian Liu\textsuperscript{1,$*$} \hspace{5mm} Xinyu Yang\textsuperscript{1,2,$*$} \hspace{5mm} Haotian Tang\textsuperscript{1} \hspace{5mm} Shang Yang\textsuperscript{1,3} \hspace{5mm} Song Han\textsuperscript{1} \\
\textsuperscript{1}MIT \hspace{10mm} \textsuperscript{2}Shanghai Jiao Tong University \hspace{10mm} \textsuperscript{3}Tsinghua University \\\\
\url{https://flatformer.mit.edu}
}

\maketitle

\footnotetext{$*$ indicates equal contributions.}

\begin{abstract}

Transformer, as an alternative to CNN, has been proven effective in many modalities (\eg, texts and images). For 3D point cloud transformers, existing efforts focus primarily on pushing their accuracy to the state-of-the-art level. However, their latency lags behind sparse convolution-based models (\textbf{3$\times$ slower}), hindering their usage in resource-constrained, latency-sensitive applications (such as autonomous driving). This inefficiency comes from point clouds' sparse and irregular nature, whereas transformers are designed for dense, regular workloads. This paper presents \textbf{\model} to close this latency gap by trading spatial proximity for better computational regularity. We first flatten the point cloud with window-based sorting and partition points into \textbf{groups of equal sizes} rather than \textbf{windows of equal shapes}. This effectively avoids expensive structuring and padding overheads. We then apply self-attention within groups to extract local features, alternate sorting axis to gather features from different directions, and shift windows to exchange features across groups. \model delivers state-of-the-art accuracy on Waymo Open Dataset with \textbf{4.6$\times$} speedup over (transformer-based) SST and \textbf{1.4$\times$} speedup over (sparse convolutional) CenterPoint. This is the first point cloud transformer that achieves real-time performance on edge GPUs and is faster than sparse convolutional methods while achieving on-par or even superior accuracy on large-scale benchmarks.

\end{abstract}
\section{Introduction}

\begin{figure}[!t]
    \centering
    \includegraphics[width=\linewidth]{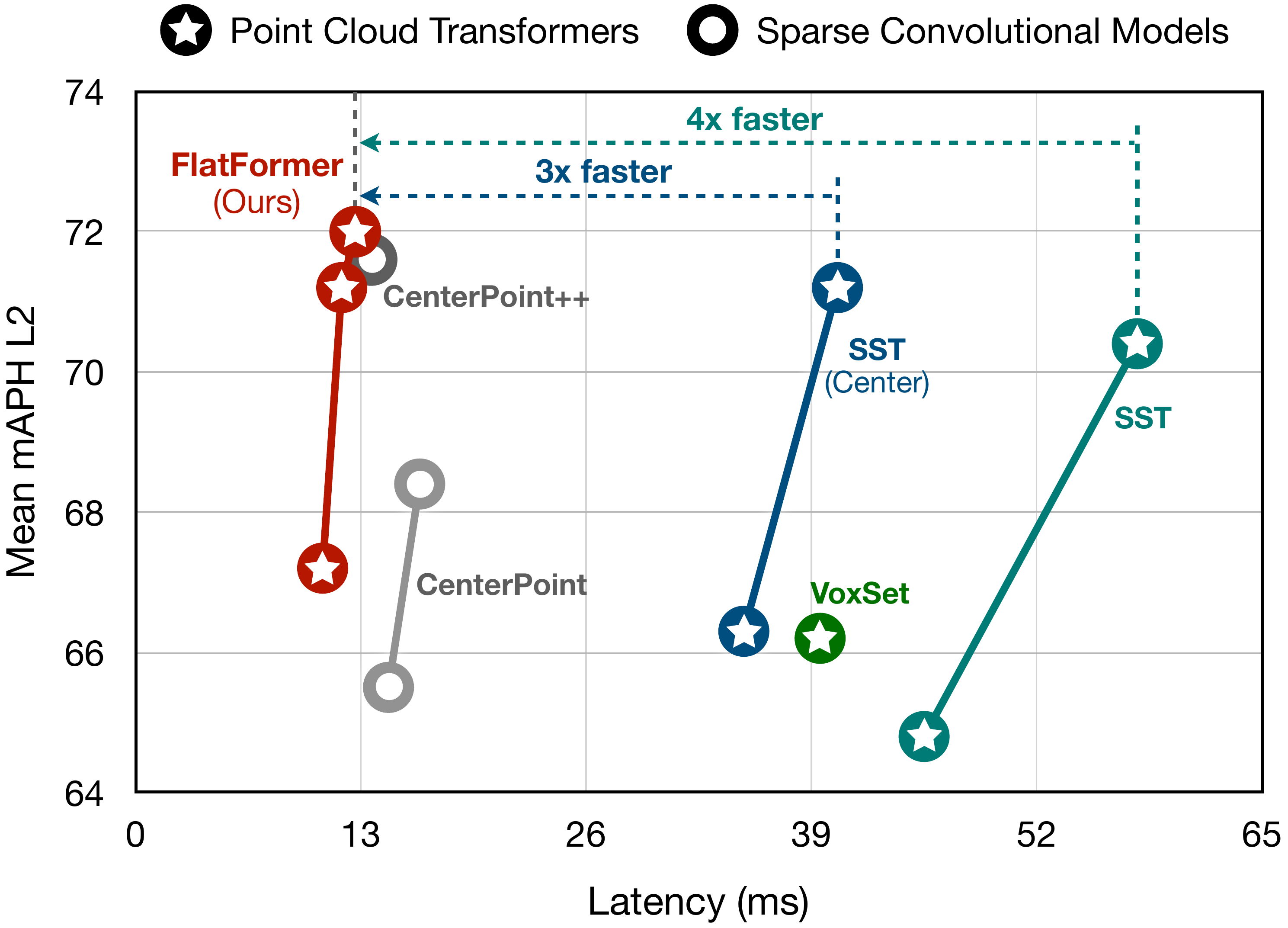}
    \caption{Previous point cloud transformers ($\star$) are \textbf{3-4$\times$} slower than sparse convolution-based models ($\bullet$) despite achieving similar detection accuracy. \model is the first point cloud transformer that is faster than sparse convolutional methods with on-par accuracy. Latency is measured on an NVIDIA Quadro RTX A6000.}
    \label{fig:teaser}
    \vspace{-8pt}
\end{figure}

Transformer~\cite{vaswani2017attention} has become the model of choice in natural language processing (NLP), serving as the backbone of many successful large language models (LLMs)~\cite{devlin2018bert,brown2020language}. Recently, its impact has further been expanded to the vision community, where vision transformers (ViTs)~\cite{dosovitskiy2020image,touvron2021training,liu2021swin} have demonstrated on-par or even superior performance compared with CNNs in many visual modalities (\eg, image and video). 3D point cloud, however, is not yet one of them.

Different from images and videos, 3D point clouds are intrinsically sparse and irregular. Most existing point cloud models~\cite{yin2021center} are based on 3D sparse convolution~\cite{graham20183d}, which computes convolution only on non-zero features. They require dedicated system support~\cite{yan2018second,choy20194d,tang2022torchsparse} to realize high utilization on parallel hardware (\eg, GPUs).

Many efforts have been made toward point cloud transformers (PCTs) to explore their potential as an alternative to sparse convolution. Global PCTs~\cite{guo2021pct} benefit from the regular computation pattern of self-attention but suffer greatly from the quadratic computational cost (\wrt the number of points). Local PCTs~\cite{zhao2021point,mao2021voxel} apply self-attention to a local neighborhood defined in a similar way to point-based models~\cite{qi2017pointnet++} and are thus bottlenecked by the expensive neighbor gathering~\cite{liu2019pvcnn}. These methods are only applicable to single objects or partial indoor scans (with $<$4k points) and cannot be efficiently scaled to outdoor scenes (with $>$30k points).

Inspired by Swin Transformer~\cite{liu2021swin}, window PCTs~\cite{fan2022embracing,sun2022swformer} compute self-attention at the window level, achieving impressive accuracy on large-scale 3D detection benchmarks. Despite being spatially regular, these windows could have drastically different numbers of points (which differ by more than \textbf{80$\times$}) due to the sparsity. This severe imbalance results in redundant computation with inefficient padding and partitioning overheads. As a result, window PCTs can be \textbf{3$\times$ slower} than sparse convolutional models (\fig{fig:teaser}), limiting their applications in resource-constrained, latency-sensitive scenarios (\eg, autonomous driving, augmented reality).

This paper introduces \textbf{\model} to close this huge latency gap. Building upon window PCTs, \model trades spatial proximity for better computational regularity by partitioning 3D point cloud into \textit{groups of equal sizes} instead of \textit{windows of equal shapes}. It applies self-attention within groups to extract local features, alternates the sorting axis to aggregate features from different orientations, and shifts windows to exchange features across groups. Benefit from the regular computation pattern, \model achieves \textbf{4.6$\times$} speedup over (transformer-based) SST and \textbf{1.4$\times$} speedup over (sparse convolutional) CenterPoint while delivering the state-of-the-art accuracy on Waymo Open Dataset.

To the best of our knowledge, \model is the first point cloud transformer that achieves on-par or superior accuracy than sparse convolutional methods with lower latency. It is also the first to achieve real-time performance on edge GPUs. With better hardware support for transformers (\eg, NVIDIA Hopper), point cloud transformers will have a huge potential to be the model of choice in 3D deep learning. We believe our work will inspire future research in this direction.
\section{Related Work}

\paragraph{Deep Learning on Point Clouds.}

Early research converts point clouds from 3D sensors to dense voxel grids and applies 3D CNNs~\cite{maturana2015voxnet,qi2016volumetric,cicek20163d,wu20153d} on the volumetric inputs. However, the compute and memory consumption of volumetric CNNs grows cubically \wrt the input resolution, limiting the scalability of these methods. To overcome this bottleneck, later research~\cite{qi2017pointnet,qi2017pointnet++,wang2018dynamic,li2018pointcnn,wu2019pointconv,thomas2019kpconv,hu2019randla,qian2021assanet} directly performs feature extraction on point sets, while \cite{riegler2017octnet,wang2017cnn,wang2018adaptive} convert point clouds to octrees and \cite{graham20183d,choy20194d,yan2018second,chen2022scaling,liu2021efficient} perform sparse convolution on sparse voxels. Recently, researchers also explore point+voxel~\cite{liu2019pvcnn,liu2021pvnas,mao2019interpolated,xu2020gridgcn} or point+sparse voxel~\cite{shi2020pv,shi2021pv,mao2019interpolated,tang2020searching} hybrid representations for efficient 3D deep learning. 

\myparagraph{3D Object Detection.}

Extensive attention has been paid to 3D object detection~\cite{geiger2013vision,caesar2020nuscenes,sun2020scalability} for autonomous vehicles. Early research~\cite{qi2017frustum,wang2019frustum} generates object proposals on 2D images and refines the predictions in the lifted 3D frustums. VoxelNet~\cite{zhou2018voxelnet} leads another line of research that directly detects 3D objects without 2D proposals. Following VoxelNet, PointPillars~\cite{lang2019pointpillars}, SECOND~\cite{yan2018second,zhu2019classbalanced}, 3DSSD~\cite{yang20203dssd} and MVF~\cite{zhou2019mvf} are all single-stage anchor-based 3D detectors, while CenterPoint~\cite{yin2021center,yin2022centerpoint++}, AFDet~\cite{ge2021afdet,ge2021afdetv2}, HotspotNet~\cite{chen2020object}, MVF++~\cite{qi2021offboard}, RangeDet~\cite{fan2021rangedet}, 
PolarStream~\cite{chen2021polarstream}, ObjectDGCNN~\cite{wang2021object}, M3DETR~\cite{guan2022m3detr}, PillarNet~\cite{shi2022pillarnet}, LidarMultiNet~\cite{ye2022lidarmultinet} are single-stage anchor-free 3D detectors. PointRCNN~\cite{shi2019pointrcnn}, Fast Point R-CNN~\cite{chen2019fast}, Part-A\textsuperscript{2}Net~\cite{shi2019part}, PV-RCNN~\cite{shi2020pv,shi2021pv}, LiDAR R-CNN~\cite{li2021lidar}, CenterFormer~\cite{zhou2022centerformer}, FSD~\cite{fan2022fully}, MPPNet~\cite{chen2022mppnet} add a second stage that refines the proposals from the region proposal network (RPN) in the 3D space. There are also recent explorations on multi-sensor 3D object detection~\cite{vora2020pointpainting,yin2021multimodal,li2022deepfusion,liu2022bevfusion,liang2018deep,bai2022transfusion,chen2022focal,chen2017mv3d}.

\myparagraph{Vision Transformers.}

Motivated by the huge success of transformers~\cite{vaswani2017attention,devlin2018bert} in natural language processing (NLP), researchers have started to adapt transformers to various vision tasks~\cite{khan2021transformers}. The pioneering ViT~\cite{kolesnikov2021vit} first demonstrates that an image can be viewed as 16$\times$16 words and processed by multi-head attention layers. DeiT~\cite{touvron2021training} further shows that ViTs can be trained in a data-efficient manner without pretraining on JFT~\cite{sun2017revisiting}. T2T-ViT~\cite{yuan2021t2tvit}, Pyramid ViT~\cite{wang2021pyramid,wang2021pvtv2} and CrossFormer~\cite{wang2021crossformer} introduce hierarchical modeling capability to ViTs. Swin Transformer~\cite{liu2021swin,liu2022swinv2} limits self-attention computation to non-overlapping windows and enables cross-window information exchange via window shifting. There are also task-specific ViTs such as ViTDet~\cite{li2022exploring} for object detection, SETR~\cite{zheng2021rethinking}, and SegFormer~\cite{xie2021segformer} for semantic segmentation. Instead of adopting a fully-transformer backbone, another line of research, such as DETR~\cite{carion2020end}, Deformable DETR~\cite{zhu2020deformable}, MaskFormer~\cite{cheng2021maskformer}, PanopticSegFormer~\cite{li2021panoptic}, DETR3D~\cite{wang2021detr3d}, BEVFormer~\cite{li2022bevformer}, apply self-attention only to the task-specific heads and still uses CNNs for the backbone.

\myparagraph{Point Cloud Transformers.}

Recently, fully-transformer architectures have begun to emerge in the point cloud domain. Similar to ViT, PCT~\cite{guo2021pct} calculates self-attention globally on the entire point cloud, which falls short in scalability as its computation complexity scales quadratically as the number of points grows. PointASNL~\cite{yan2020pointasnl}, PointTransformer~\cite{zhao2021point,wu2022point}, Fast Point Transformer~\cite{park2022fast}, PointFormer~\cite{pan20213d}, VoTr~\cite{mao2021voxel}, VoxSet~\cite{he2022voxel} applies transformer-based architecture on the local neighborhood of each point. The efficiency of these local transformers is limited by neighborhood query and feature restructuring. Most related to our work are the window-based point cloud transformers, SST~\cite{fan2022embracing} and SWFormer~\cite{sun2022swformer}. Inspired by Swin Transformer, they project the point cloud into a bird's-eye view and divide the BEV space into non-overlapping windows with the same \textit{spatial sizes} (but different \textit{number of points}). Window shifting is used to communicate information across windows. SST suffers from large computation in window partition and padding overhead due to regional grouping, and achieves only \textbf{one-sixth} utilization compared with sparse convolutional models.  

In this paper, we only refer to those methods that adopt a transformer-based architecture in the \textit{backbone} as point cloud transformers. As CenterFormer~\cite{zhou2022centerformer}, FUTR3D~\cite{chen2022futr3d} and UVTR~\cite{li2022uvtr} apply sparse convolutional backbones and only use the transformer in their detection heads, we still categorize them as sparse convolutional methods.

\section{Why are Point Cloud Transformers Slow?}
\label{sect:background}

Although point cloud transformers (PCTs) start to catch up with the accuracy of sparse convolutional detectors, there is still a 3$\times$ latency gap between the fastest PCT (SST~\cite{fan2022embracing}) and sparse convolutional CenterPoint~\cite{yin2021center} (\fig{fig:teaser}). In this section, we dissect the efficiency bottleneck of PCTs, which lays a solid foundation for our \model design. 

\subsection{Global PCTs}

\begin{figure}[!h]
    \centering
    \includegraphics[width=\linewidth]{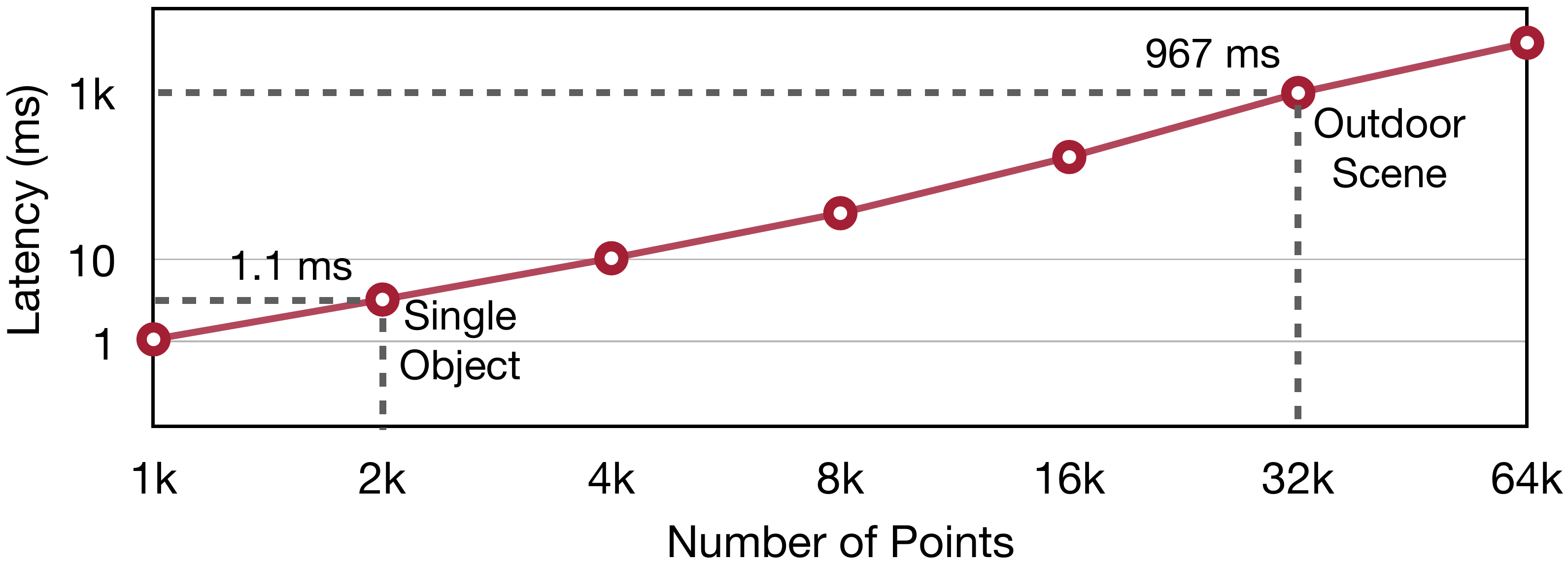}
    \caption{Latency of global PCTs scale quadratically with respect to the number of input points and cannot scale up to outdoor scenes.}
    \label{fig:background:global}
    \vspace{-8pt}
\end{figure}

Inspired by ViT~\cite{kolesnikov2021vit}, the most simple and straightforward design for transformers on point cloud is global PCTs~\cite{guo2021pct}, where each point is a token. They leverage multi-head self-attention (MHSA)~\cite{vaswani2017attention} globally across the entire point cloud. While being effective on small-scale 3D objects, global PCTs fall short in scaling to large-scale scenes due to its $\mathcal{O}(N^2D)$ complexity, where $N$ is the number of tokens and $D$ is the number of channels. From \fig{fig:background:global}, the runtime of global PCTs~\cite{guo2021pct} grows quadratically as the number of input points grows. For example, with 32k input points\footnote{32k is the number of points left after 0.32m$\times$0.32m BEV projection in a single-frame Waymo~\cite{sun2020scalability} scene.}, the model takes \textbf{almost one second} to execute on an NVIDIA A6000 GPU, \textbf{66$\times$} slower than CenterPoint~\cite{yin2021center}.

\subsection{Local PCTs}

\begin{figure}[!h]
    \centering
    \includegraphics[width=\linewidth]{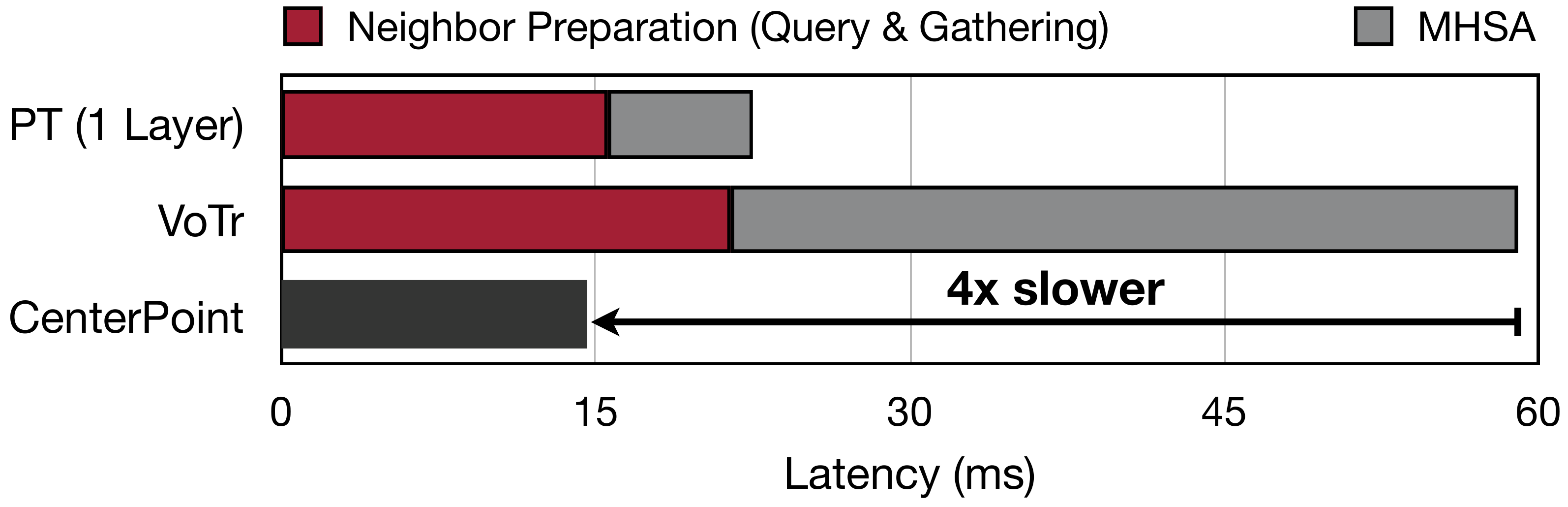}
    \caption{Local PCTs suffer from large neighborhood query and data restructuring overhead.}
    \label{fig:background:local}
    \vspace{-8pt}
\end{figure}

Researchers have proposed local PCTs~\cite{yan2020pointasnl,zhao2021point,wu2022point,park2022fast,pan20213d,mao2021voxel} to solve the scalability issue of global PCTs. They apply MHSAs to the neighborhood of each point rather than the entire point cloud. Hence, their computational complexity is $\mathcal{O}(NK^2D)$, where $N$ is the number of points, $K$ is the number of neighbors for each point, and $D$ is the number of channels. As $N$ ranges from 20k to 100k for real workloads and $K$ is less than 64 for local PCTs, their theoretical cost is much lower than global PCTs.

Local PCTs, however, suffer greatly from neighbor preparation overheads. As point cloud is sparse and irregular, we have to first \textit{find the neighbors} of each point, and then re-\textit{restructure the data} from the $N\times D$ format to the $N\times K\times D$ format on which MHSAs can be applied. These two steps are slow, taking \textbf{22} ms (\ie, \textbf{36\%} of the total runtime) for VoTr~\cite{mao2021voxel} to execute for a single scene on Waymo, which is already slower than the entire CenterPoint model. For Point Transformer (PT)~\cite{zhao2021point}, the cost for preparing neighbors takes up to \textbf{70\%} of the runtime. Such overhead in a \textit{single} layer is already larger than the total runtime of CenterPoint!

\subsection{Window PCTs}

\begin{figure}[!h]
    \centering
    \includegraphics[width=\linewidth]{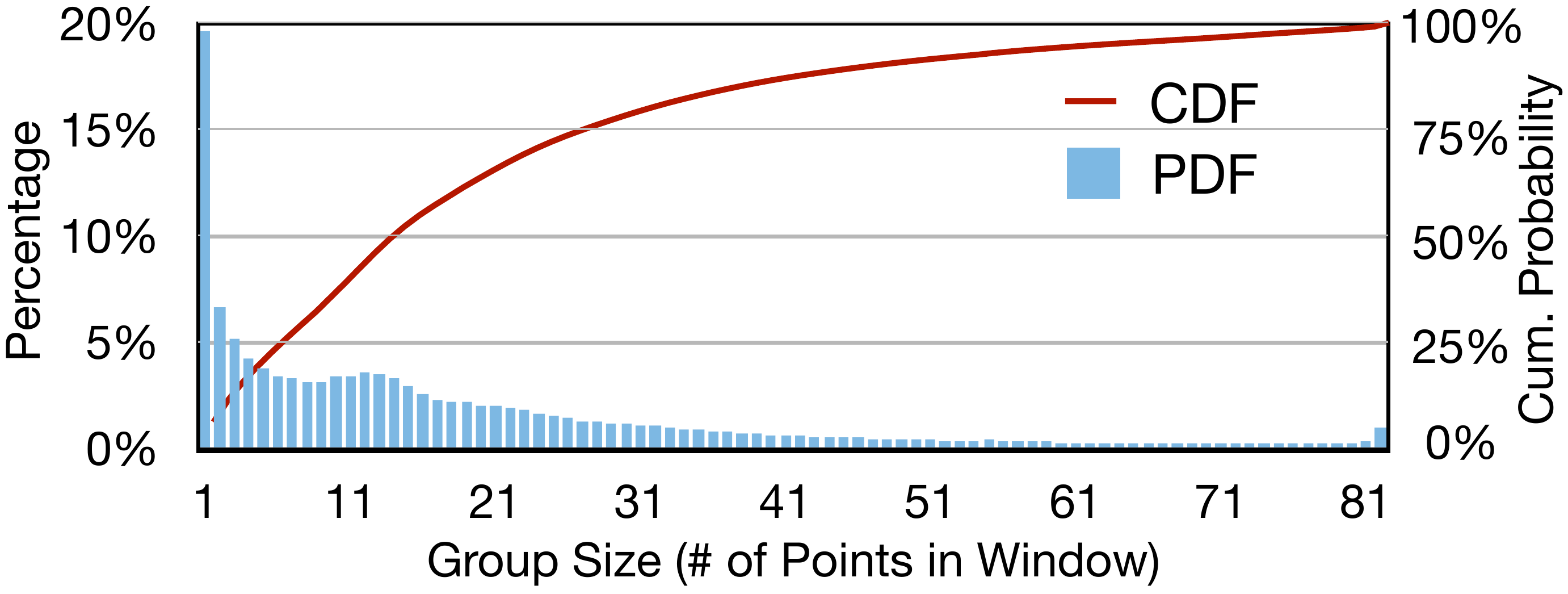}
    \caption{In SST~\cite{fan2022embracing}, the number of points within each window has a large variance. Therefore, padding is necessary and leads to significant overhead for MHSA computation.}
    \label{fig:background:window}
    \vspace{-8pt}
\end{figure}

\begin{figure*}[!t]
    \centering
    \includegraphics[width=\linewidth]{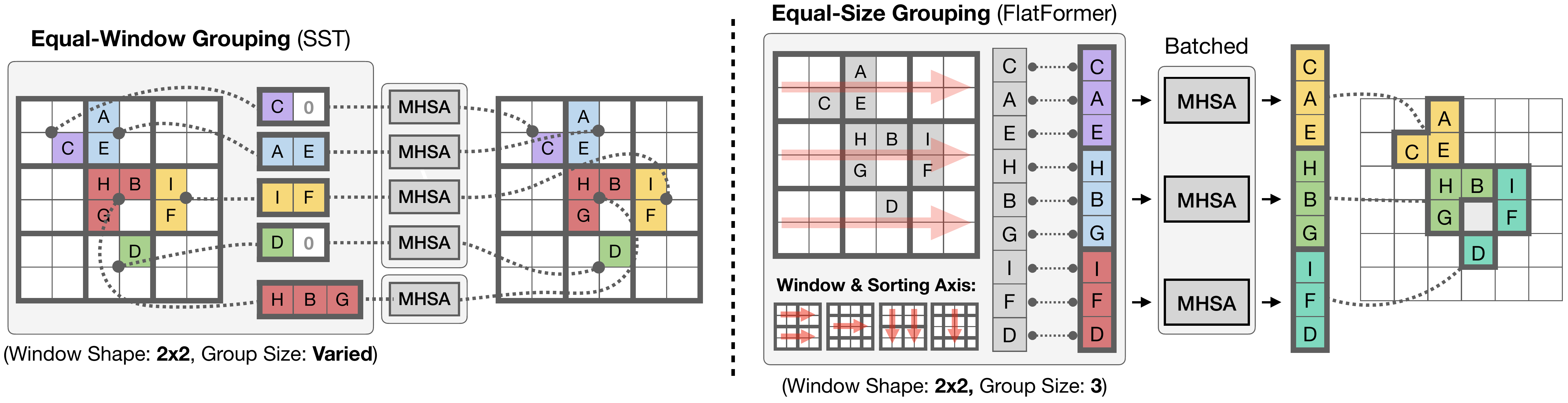}
    \caption{\model partitions the point cloud into groups of equal sizes (\textit{right}), rather than windows of equal shapes (\textit{left}). This effectively trades \textit{spatial proximity} for better \textit{computational regularity}. It then applies self-attention within each group to extract local features, alternates the sorting axis to aggregate features from different directions, and shifts windows to exchange features across groups.}
    \label{fig:overview}
    \vspace{-8pt}
\end{figure*}

The great success of Swin Transformers~\cite{liu2021swin,liu2022swinv2} in various visual recognition tasks motivates the design of window PCTs, among which, SST~\cite{fan2022embracing} is a representative work. It first projects the point cloud into the bird's-eye-view (BEV) space, then divides the BEV space into equally-shaped, non-overlapping windows, and applies MHSA within each window. Similar to Swin Transformer, SST uses window shifting to enable information exchange across windows.

Different from images, point clouds are sparse and non-uniformly distributed over the space. As a result, the number of point within each window is not the same and can differ by two orders of magnitude (\fig{fig:background:window}). As the vanilla MHSA kernel cannot efficiently support variable sequence lengths, SST~\cite{fan2022embracing} batches windows with similar sizes together and pad all windows in each batch to the largest group size within the batch (\fig{fig:overview}\textcolor{red}{l}). It then applies MHSA within each batch separately. In practice, such padding introduces a \textbf{1.7$\times$} computation overhead on Waymo. Worse still, partitioning points to equal windows also introduce significant latency overhead: it takes \textbf{18 ms} per scene on Waymo, even slower than the total runtime of CenterPoint. To sum up, the padding and partitioning overheads make SST less hardware-friendly compared with sparse convolutional methods.

\section{\model}

With all the lessons learned in \sect{sect:background}, we will design our point cloud transformer to be scalable and efficient.

\subsection{Overview}

The basic building block of \model is Flattened Window Attention (FWA). As in \fig{fig:overview}\textcolor{red}{r}, FWA adopts \emph{window-based sorting} to flatten the point cloud and partitions it to \textit{groups of equal sizes} rather than \textit{windows of equal shapes}. This naturally resolves the group size imbalance problem and avoids the padding and partitioning overheads. FWA then applies self-attention within groups to extract local features, alternates sorting axis to aggregate features from different orientations, and shifts windows to exchange features across groups. Finally, we provide an implementation of FWA that further improves its efficiency and minimizes the overheads.

\subsection{Flattened Window Attention (FWA)}

\subsubsection{Sorting \& Grouping}

\paragraph{Window-Based Sorting.}

With a point cloud $\{(x, y)\}$\footnote{We assume that the point cloud is in 2D for ease of notation, while our method applies to 3D or higher-dimension point clouds.}, we first quantize the coordinate of each point $(x, y)$ to
\begin{equation}
    \bigl(\,\underbrace{\lfloor x / w_x \rfloor, \lfloor y / w_y \rfloor}_{\text{window coordinates}},\,\,\underbrace{x - \lfloor x / w_x \rfloor \cdot w_x, y - \lfloor y / w_y \rfloor \cdot w_y}_{\text{local coordinates within window}}\,\bigr),
\end{equation}
where $(w_x, w_y)$ is the window shape. Next, we sort all points first by \textit{window coordinates} and then by \textit{local coordinates} within the window. This step turns the unordered point cloud into an ordered one, where points within the same window will be next to each other.

\myparagraph{Equal-Size Grouping.}

Conventional window PCTs~\cite{fan2022embracing} will then group the points within the same window together. However, as discussed in \sect{sect:background}, each group can have drastically different numbers of points due to inherent sparsity. To overcome the padding overheads, we partition the point cloud into \textit{groups of equal sizes} based on the sorted sequence. This step allows the subsequent group attention to enjoy a perfectly regular workload. From the implementation perspective, our grouping only involves a simple tensor reshaping (which is free since it does not change the layout) and is more efficient than window partitioning in SST~\cite{fan2022embracing}.

\myparagraph{Alternate Sorting Axis.}

Between the two axes, $x$ has a higher priority in sorting. Thus, points with identical $\lfloor x / w_x \rfloor$ will be next to each other while points with the same $\lfloor y / w_y \rfloor$ can be very far away from each other in the sorted sequence, breaking the geometric locality. To solve this inequity, we alternate the sorting axis between $x$ and $y$ in different FWA blocks. This is very similar to spatially separable convolution that decomposes a 3$\times$3 kernel into 3$\times$1 and 1$\times$3 kernels. Stacking FWA blocks with different sorting axes enables the model to aggregate local features from different directions.

\myparagraph{Equal Size \vs Equal Window.}

The key design choice we made is to partition the point cloud into groups of equal sizes rather than windows of equal shapes. There is a trade-off: equal-window grouping maintains perfect \textit{spatial proximity} (\ie, each group has the same radius) but breaks the \textit{computational regularity}, while equal-size grouping ensures balanced computation workload (\ie, each group has the same number of points) but cannot guarantee the geometric locality. We show in \sect{sect:analysis} that computation regularity is more important since spatial irregularity can be partially addressed by our algorithm design: \ie, window-based sorting offers a fairly good spatial ordering, and self-attention is robust to outliers (\ie, distant point pairs).

\subsubsection{Group Attention}

With points partitioned, we then apply self-attention~\cite{vaswani2017attention} within each group to extract local features. For each group of points with coordinates $\mathcal{C}$ and features $\mathcal{F}$, we follow the standard transformer block design:
\begin{equation}
\begin{aligned}
    \mathcal{F}' &= \mathcal{F} + \texttt{MHSA}(\texttt{LN}(\mathcal{F}), \texttt{PE}(\mathcal{C})), \\
    \mathcal{F}'' &= \mathcal{F}' + \texttt{FFN}(\texttt{LN}(\mathcal{F}')),
\end{aligned}
\end{equation}
where $\texttt{MHSA}(\cdot)$, $\texttt{FFN}(\cdot)$ and $\texttt{LN}(\cdot)$ correspond to multi-head self-attention, feed-forward layer, and layer normalization, respectively. Different from SST~\cite{fan2022embracing}, $\texttt{PE}(\cdot)$ gives global absolute positional embedding. Here, we use the most standard \texttt{softmax} attention formulation for $\texttt{MHSA}(\cdot)$. Our method will benefit from other more efficient attention variants, such as linear attention~\cite{katharopoulos2020transformers}, which we leave for future work.

\myparagraph{Window Shift.}

Benefit from the non-overlapping design, window-based attention typically has a larger receptive field than convolution (\eg, 69 neighbors in our FWA \vs $\leq$27 neighbors in a sparse convolution of kernel size 3). However, its modeling power is limited as there is no feature exchange across groups. Similar to Swin Transformer~\cite{liu2021swin,liu2022swinv2,fan2022embracing}, we adopt the shifted window approach that alternates the sorting configuration in consecutive FWA blocks. Specifically, we translate the coordinates of all points by $(w_x/2, w_y/2)$ for sorting in shifted FWA blocks. This mechanism introduces cross-group feature communication while effectively maintaining workload independence between groups. Note that alternating sorting axis also enables feature exchange.

\subsection{Efficient Implementation}
\label{sect:model:system}

Besides the algorithm design, we also provide an implementation that improves the efficiency of MHSA and FFN and minimizes the sorting and masking overheads. All these optimizations are specialized for our point cloud transformer design and are not applicable to sparse convolution models.

\myparagraph{Efficient MHSA.}

Within MHSA, query $\mathcal{Q}$, key $\mathcal{K}$, and value $\mathcal{V}$ will first be transformed with separate linear layers. We pack these three linear projections into a batched matrix multiplication (since $\mathcal{Q}$, $\mathcal{K}$ and $\mathcal{V}$ have the same shape in our FWA) to improve the parallelism. In addition, standard attention implementations materialize $\mathcal{Q}\mathcal{K}^\text{T}$ and $\texttt{softmax}(\mathcal{Q}\mathcal{K}^\text{T})$. We leverage a recent efficient functional-preserving implementation (FlashAttention~\cite{dao2022flashattention}) that uses tiling to reduce the number of memory reads/writes, achieving better efficiency.

\myparagraph{Efficient FFN.}
 
FFN consists of two linear layers with a \texttt{GELU} activation in the middle. We implement a fused linear kernel (in \texttt{Triton}) that absorbs the activation into the layer before to avoid writing the intermediate results to DRAM. We also observe that our linear kernel (optimized by \texttt{Triton}) is even more efficient than \texttt{cuBLAS}, which is probably due to the unconventional tall-and-skinny workload.

\myparagraph{Reuse Sorting.}

Sorting the coordinates of all points is a non-negligible overhead. As the coordinates remain identical (w/o downsampling), we reuse the sorting results (\ie, ranks of each point) with the same axis and window. In practice, this reduces the sorting overhead in our model by 50\%.

\myparagraph{Drop Residual.}

The size of the input point cloud might not be divisible by the group size, generating a group with fewer points after partition. This minor irregularity will still result in some overheads in self-attention since we need to introduce masking to correctly zero them out. Instead, we directly drop the final non-full group. This only corresponds to less than 0.1\% of all points, having a negligible impact on the model's performance ($<$0.1\%).

\section{Experiments}
\label{sect:exp}

\begin{table*}[!t]
    \small\centering
    \begin{tabular}{lcccccccc}
        \toprule
        & \multirow{2}{*}{\#Frames} & \#MACs & Latency & Speedup & Mean L2 & Vehicle L2 & Pedestrian L2 & Cyclist L2 \\
        & & (G) & (ms) & (\wrt \cite{yin2021center}) & (mAPH) & (mAP/APH) & (mAP/APH) & (mAP/APH) \\
        \midrule
        \textcolor{darkgray!50}{SECOND~\cite{yan2018second}\textsuperscript{1}} & \textcolor{darkgray!50}{1} & \textcolor{darkgray!50}{--} & \textcolor{darkgray!50}{--} & \textcolor{darkgray!50}{--} & \textcolor{darkgray!50}{57.2} & \textcolor{darkgray!50}{63.9 / 63.3} & \textcolor{darkgray!50}{60.7 / 51.3} & \textcolor{darkgray!50}{58.3 / 57.0} \\
        \textcolor{darkgray!50}{PointPillars~\cite{lang2019pointpillars}\textsuperscript{1}} & \textcolor{darkgray!50}{1} & \textcolor{darkgray!50}{--} & \textcolor{darkgray!50}{--} & \textcolor{darkgray!50}{--} & \textcolor{darkgray!50}{57.8} & \textcolor{darkgray!50}{63.6 / 63.1} & \textcolor{darkgray!50}{62.8 / 50.3} & \textcolor{darkgray!50}{61.9 / 59.9} \\
        \cdashlinelr{1-9}
        \spc CenterPoint~\cite{yin2021center}\textsuperscript{2} & 1 & 126.9 & 14.6 & 1.0$\times$ & 65.5 & 66.7 / 66.2 & 68.3 / 62.6 & 68.7 / 67.6 \\
        \pct VoTr-SSD~\cite{mao2021voxel} & 1 & 110.3 & \,\,\,59.1$^\dag$ & 0.2$\times$ & -- & 60.2 / 59.7 & -- & -- \\
        \pct SST~\cite{fan2022embracing}\textsuperscript{3} & 1 & 204.9 & 45.5 & 0.3$\times$ & 64.8 & 64.8 / 64.4 & 71.7 / 63.0 & 68.0 / 66.9 \\
        \pct SST-Center~\cite{fan2022embracing} & 1 & 226.4 & 35.1 & 0.4$\times$ & 66.3 & 66.6 / 66.2 & 72.4 / 65.0 & 68.9 / 67.6 \\
        \pct VoxSet~\cite{he2022voxel} & 1 & 189.4 & 39.5 & 0.4$\times$ & 66.2 & 66.0 / 65.6 & 72.5 / 65.4 & 69.0 / 67.7 \\
        \spc PillarNet~\cite{shi2022pillarnet} & 1 & 138.3 & 11.1 & 1.3$\times$ & \textbf{67.2} & 70.4 / 69.9 & 71.6 / 64.9 & 67.8 / 66.7 \\
        \pct \textbf{\model} (Ours) & 1 & 177.2 & \textbf{10.8} & \textbf{1.4$\times$} & \textbf{67.2} & 69.0 / 68.6 & 71.5 / 65.3 & 68.6 / 67.5 \\
        \midrule
        \spc CenterPoint~\cite{yin2021center}\textsuperscript{2} & 2 & 137.6 & 16.4 & 1.0$\times$ & 68.4 & 67.7 / 67.2 & 71.0 / 67.5 & 71.5 / 70.5\\
        \spc PillarNet~\cite{shi2022pillarnet} & 2 & 148.8 & \textbf{11.6} & \textbf{1.4$\times$} & 70.0 & 71.6 / 71.1 & 74.5 / 71.4 & 68.3 / 67.5 \\
        \pct \textbf{\model} (Ours) & 2 & 186.6 & 11.9 & \textbf{1.4$\times$} & \textbf{71.2} & 70.8 / 70.3 & 73.8 / 70.5 & 73.6 / 72.6 \\
        \midrule
        \spc CenterPoint~\cite{yin2021center} & 3 & 144.7 & 18.3 & 1.0$\times$ & -- & -- & -- & -- \\
        \spc CenterPoint++~\cite{yin2022centerpoint++}\textsuperscript{2} & 3 & 113.0 & 13.6 &  1.3$\times$ & 71.6 & 71.8 / 71.4 & 73.5 / 70.8 & 73.7 / 72.8 \\
        \pct SST~\cite{fan2022embracing}\textsuperscript{3} & 3 & 250.0 & 57.8 & 0.3$\times$ & 70.4 & 66.5 / 66.1 & 76.2 / 72.3 & 73.6 / 72.8 \\
        \pct SST-Center~\cite{fan2022embracing}\textsuperscript{4} & 3 & 243.1 & 40.5 & 0.5$\times$ & 71.2 & 68.8 / 68.2 & 75.8 / 71.8 & 74.4 / 73.3 \\
        \pct SWFormer~\cite{sun2022swformer} & 3 & -- & \,\;20.0$^\ddag$ & 0.9$\times$ & -- & 71.1 / 70.6 & 74.8 / 71.1 & -- \\
        \pct \textbf{\model} (Ours) & 3 & 193.2 & \textbf{12.7} & \textbf{1.4$\times$} & \textbf{72.0} & 71.4 / 71.0 & 74.5 / 71.3 & 74.7 / 73.7 \\
        \bottomrule
    \end{tabular}
    \caption{Results of single-stage 3D detectors on Waymo Open Dataset (validation set). \model achieves 1.4$\times$ speedup over CenterPoint and 4.6$\times$ speedup over SST while being more accurate. We refer the readers to the appendix for detailed metrics (\eg, L1 mAP/mAPH). Markers \spc and \pct refer to sparse convolutional models and point cloud transformers, respectively. Methods with $<$60 L2 mAPH are marked \textcolor{darkgray!50}{gray}. (\textsuperscript{1}: from FSD paper, \textsuperscript{2}: from CenterPoint authors, \textsuperscript{3}: from SST authors, \textsuperscript{4}: reproduced by us, \textsuperscript{\dag}: projected latency, \textsuperscript{\ddag}: latency on T4)}
    \label{tab:results:1s}
    \vspace{-8pt}
\end{table*}
\begin{table}[!t]
    \setlength{\tabcolsep}{5pt}
    \small\centering
    \begin{tabular}{lcccccccc}
        \toprule
        & \multirow{2}{*}{\#Frames} & Latency & Mean L2 \\
        & & (ms) & (mAPH) \\
        \midrule
        \spc LiDAR R-CNN~\cite{li2021lidar}\textsuperscript{\dag} & 1 & -- & 61.3 \\
        \spc PV-RCNN~\cite{shi2020pv}\textsuperscript{\dag} & 1 & -- & 63.3 \\
        \spc Part-A\textsuperscript{2}~\cite{shi2020points}\textsuperscript{\dag} & 1 & -- & 63.8 \\
        \spc PV-RCNN++~\cite{shi2021pv}\textsuperscript{\dag} & 1 & -- & 64.9 \\
        \spc CenterFormer~\cite{zhou2022centerformer} & 1 & 33.8 & 69.0 \\
        \spc FSD-SpConv~\cite{fan2022fully} & 1 & 47.8 & \textbf{70.8} \\
        \pct \textbf{\model}+FSD (Ours) & 1 & 39.3 & 70.5 \\
        \midrule
        \spc CenterFormer~\cite{zhou2022centerformer} & 2 & 53.5 & 72.8 \\
        \pct \textbf{\model}+FSD (Ours) & 2 & 51.8 & \textbf{73.8} \\
        \midrule
        \spc CenterFormer~\cite{zhou2022centerformer} & 4 & 85.8 & 73.2 \\
        \spc MPPNet~\cite{chen2022mppnet} & 4 & -- & 74.2 \\
        \pct \textbf{\model}+FSD (Ours) & 3 & 60.6 & \textbf{74.8} \\
        \bottomrule
    \end{tabular}
    \caption{Results of two-stage 3D detectors on Waymo Open Dataset (validation set). \model achieves on-par or even higher accuracy compared with sparse convolutional two-stage detectors. We refer the readers to the appendix for detailed metrics (\eg, per-class L1/L2 mAP/mAPH). Markers \spc and \pct refer to SpConv-based models and point cloud transformers, respectively. (\textsuperscript{\dag}: from FSD paper)}
    \label{tab:results:2s}
    \vspace{-8pt}
\end{table}

\subsection{Setup}

\paragraph{Dataset.}

We carry out our experiments on the large-scale Waymo Open Dataset (WOD)~\cite{sun2020scalability} with 1150 LiDAR point cloud sequences. Each sequence has 200 frames, collected by a 360$^\circ$ FoV LiDAR sensor at 10 frames per second. There are four foreground classes, three of which (vehicles, pedestrians and cyclists) are used for detection metric evaluation. 

\myparagraph{Metrics.}

We follow the official metrics on Waymo to calculate the standard 3D mAP and heading-weighted 3D mAP (mAPH) of all methods. The matching IoU thresholds for vehicle, pedestrian and cyclist are set to default values (0.7, 0.5 and 0.5). Objects are divided into two difficulty levels, where objects with fewer than five laser points or annotated as hard are categorized into level 2 (L2) and other objects are defined as level 1 (L1). We mainly report L2 metrics in the main paper and provide detailed metrics in the appendix. 

\myparagraph{Model.}

Based on FWA, we provide an instantiation of \model for 3D object detection. We follow the design of PointPillars~\cite{lang2019pointpillars} to first voxelize the point cloud into sparse BEV pillars (with MLPs) at a resolution of 0.32m$\times$0.32m. We then apply eight consecutive FWA blocks with alternating sorting axes (\ie, $x$ or $y$) and window shifting configurations (\ie, on or off). All FWA blocks have a window shape of 9$\times$9 and a group size of 69. Following SST~\cite{fan2022embracing}, we do not apply any spatial downsampling, which is beneficial for small objects. Finally, we apply regular BEV encoder and a center-based detection head following CenterPoint~\cite{yin2021center,yin2022centerpoint++}.

\subsection{Main Results}

\subsubsection{Single-Stage Detectors}

\paragraph{Baseline.}

We compare our \model with state-of-the-art sparse convolutional~\cite{yin2021center,yin2022centerpoint++,shi2022pillarnet} and transformer-based~\cite{mao2021voxel,fan2022embracing,he2022voxel} single-stage 3D detectors. All models apply anchor- or center-based detection heads~\cite{yan2018second,yin2021center,yin2022centerpoint++}. We compare models with different numbers of input frames separately. 

\myparagraph{Latency.}

We measure the latency on an NVIDIA Quadro RTX A6000 GPU using FP16 precision. We adopt SpConv v2.2.3~\cite{yan2018second}, the state-of-the-art 3D sparse convolution library, to execute the 3D encoder of all sparse convolutional detectors. For transformer-based detectors, we use their official implementation to measure the runtime. All modules after the 3D encoder (\eg, BEV encoder and detection head) are executed with TensorRT 8.4. We execute all the methods on the first 1,000 samples for 50 runs (with 10 warmup runs). We report the average latency (with outliers excluded). We do not include the data loading and post-processing time.

\myparagraph{Results.}

As in \tab{tab:results:1s}, our \model achieves consistent performance improvements over both sparse convolutional and transformer-based detectors with much better efficiency. For one-frame models, \model is \textbf{4.2$\times$}, \textbf{3.3$\times$} and \textbf{3.7$\times$} faster than SST, SST-Center and the recent VoxSet~\cite{he2022voxel}. It also compares favorably with strong sparse convolutional baselines: \textbf{1.4$\times$} faster than CenterPoint with 1.7 higher L2 mAPH and performs on par with PillarNet~\cite{shi2022pillarnet}. The accuracy advantage magnifies in the two-frame setting. Specifically, \model is 1.4$\times$ faster than CenterPoint with 2.8 L2 mAPH higher accuracy, and outperforms PillarNet by 1.3 L2 mAPH with a similar latency. With three input frames, \model is \textbf{4.6$\times$} and \textbf{3.2$\times$} faster than SST and SST-Center, respectively. It also achieves better latency-accuracy tradeoff (1.1$\times$ faster and 0.4\% higher accuracy) compared with CenterPoint++~\cite{yin2022centerpoint++}. Remarkably, \model requires 1.7$\times$ more MACs than CenterPoint++ while it is still faster. This indicates that our design is more hardware-friendly than the sparse convolutional baselines. 

\myparagraph{Deployment.}

\begin{figure}[!t]
    \centering
    \includegraphics[width=\linewidth]{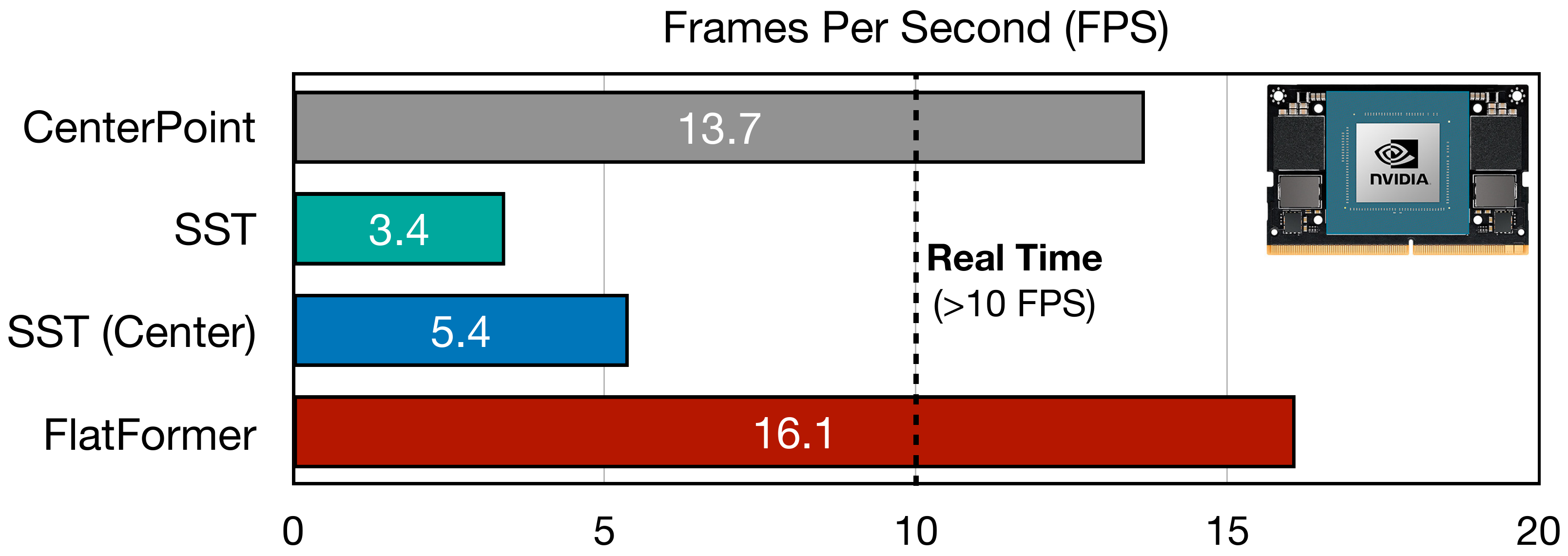}
    \caption{Measured latency on NVIDIA Jetson AGX Orin. \model is the first point cloud transformer that achieves real-time performance on edge GPUs.}
    \label{fig:results:orin}
    \vspace{-8pt}
\end{figure}

We deploy our \model on an NVIDIA Jetson AGX Orin. This is a resource-constrained edge GPU platform that is widely used in real-world self-driving cars. From \fig{fig:results:orin}, \model runs at 16 FPS, which is \textbf{1.2$\times$} faster than CenterPoint~\cite{yin2021center} and \textbf{3$\times$} faster than SST-Center. To the best of our knowledge, \model is the first point cloud transformer that achieves real-time inference (\ie, $>$10 FPS, which is the LiDAR sensor frequency) on edge GPUs. We believe that it paves the way for efficient LiDAR-centric perception in real-world applications. 

\subsubsection{Two-Stage Detectors}

\paragraph{Model.}

To verify the generalizability, we replace the 3D backbone in FSD~\cite{fan2022fully}, a state-of-the-art two-stage detector, and compare its results with previous two-stage models. We keep the same grid resolution, window shape and group size as in our single-stage experiments. 

\myparagraph{Baseline \& Latency.}

We compare our model against state-of-the-art two-stage detectors in \tab{tab:results:2s}. We follow the same latency measurement protocol. For CenterFormer~\cite{zhou2022centerformer}, we adapt the official implementation to support SpConv v2.2.3 backend in FP16 precision for a fair comparison. 

\myparagraph{Results.}

All existing high-performing two-stage detectors are sparse convolutional, while our \model is the only transformer-based method that achieves state-of-the-art level accuracy. It also shows better scalability with respect to the number of input frames compared with CenterFormer~\cite{zhou2022centerformer}. Note that our paper focuses on optimizing the latency of \textit{3D backbone}. However, two-stage detectors~\cite{zhou2022centerformer,fan2022fully} are usually bottlenecked by \textit{the second stage} in runtime, which is out of our scope. We expect that the latency of \model could benefit from a more efficient second-stage design. 
\subsection{Analysis}
\label{sect:analysis}

In this section, we present analyses to validate the effectiveness of our design choices. All experiments are based on our single-frame model trained with 20\% data.

\subsubsection{Flattened Window Attention}

\begin{figure}[!t]
    \centering
    \includegraphics[width=\linewidth]{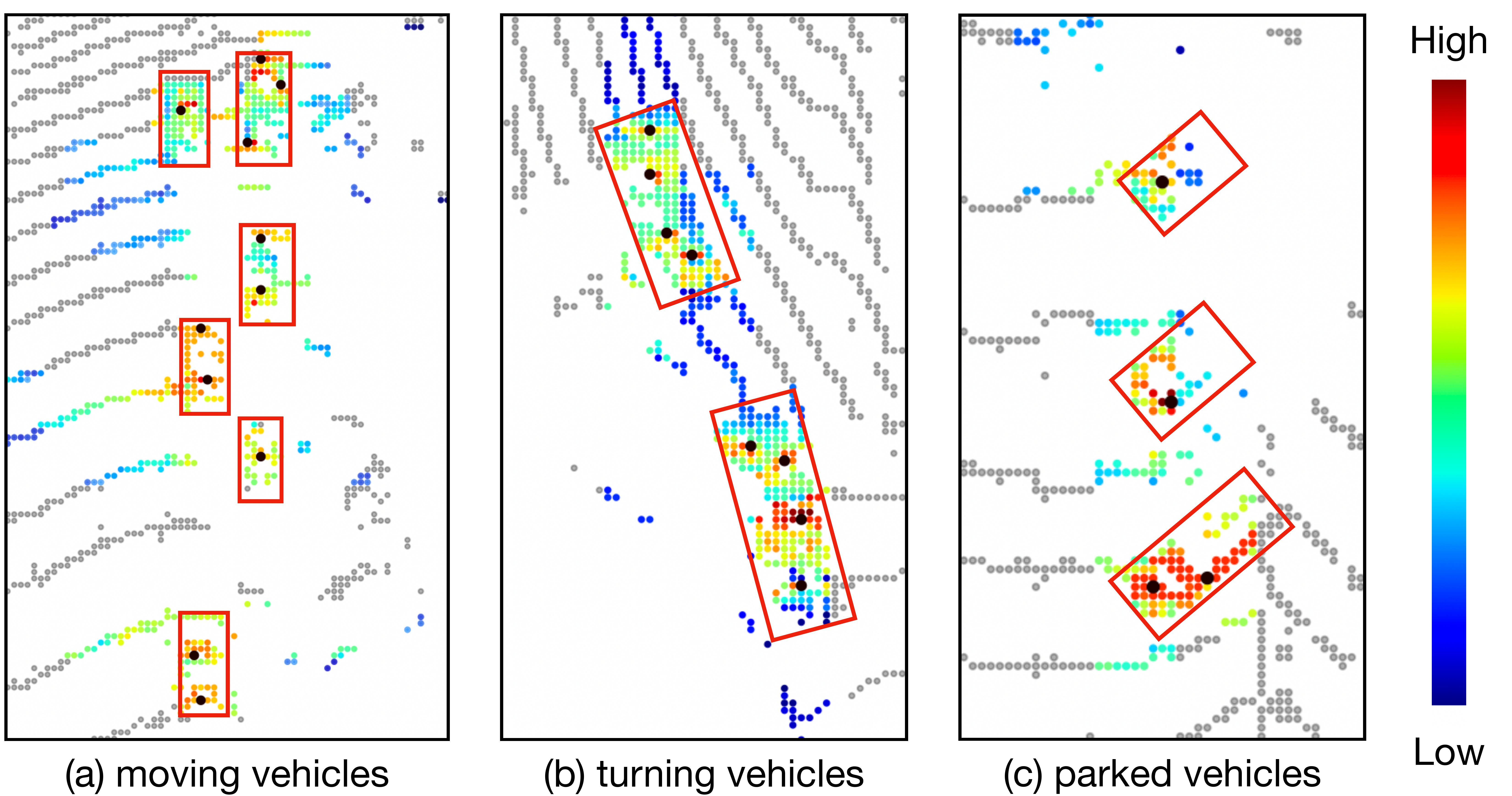}
    \caption{Visualization of attention weights in \model for vehicles that are moving straight ahead, turning and parking. High attention weights corresponds to the detected object.}
    \label{fig:analysis:visualization}
\end{figure}

In \fig{fig:analysis:visualization}, we visualize the learned attention weights in our FWA. The color represents the scale of attention weights, where warmer color means larger attention weights. Black points correspond to query points, and gray points are those with weights smaller than a threshold. For vehicles moving straight ahead, turning and parked, query points on the vehicle are always highly attended to nearby points on the same car, while faraway points have very small learned attention weights. Such an observation can partially explain the effectiveness of FWA: \ie, even if equal-size grouping does not create spatially regular windows, the model can learn to suppress the importance of outlier points in the background and focus on important foreground points within each group.

\subsubsection{Ablation Studies on Model Design}

\paragraph{Sorting Strategy.}

\begin{table}[!t]
    \setlength{\tabcolsep}{3pt}
    \small\centering
    \begin{tabular}{lcccc}
        \toprule
        \multirow{2}{*}{Sorting Strategy} & Mean L2 & Veh. L2 & Ped. L2 & Cyc. L2 \\
        & (mAPH) & (mAPH) & (mAPH) & (mAPH) \\
        \midrule
        Ours & \textbf{61.7} & \textbf{63.3} & \textbf{57.9} & \textbf{63.9}\\
        \midrule
        \textit{w/o} Quantization & 60.4 & 61.9 & 57.6 & 61.7 \\
        \textit{w/o} Axis Alternation & 61.1 & 63.1 & 57.3 & 62.9 \\
        \textit{w/o} Window Shift & 61.2 & 62.9 & 57.8 & 62.8 \\
        \midrule
        Random Order & 57.8 & 58.8 & 55.1 & 59.4 \\
        SST~\cite{fan2022embracing}  & 60.7 & 62.3 &  56.7 &  63.1 \\ 
        \bottomrule
    \end{tabular}
    \caption{Window-based sorting in \model provides even better performance than equally-shaped window partition in SST~\cite{fan2022embracing} and outperforms other sorting strategies.}
    \label{tab:analysis:sorting}
    \vspace{-8pt}
\end{table}

We first analyze the effectiveness of our window-based, axis-alternating sorting strategy in \tab{tab:analysis:sorting}. Randomly grouping points together without any spatial sorting will give $\sim$ 4\% worse performance compared with \model. Furthermore, due to spatial discontinuities on the boundary regions, directly sorting the points by $xyz$ coordinates or window sorting along a single axis both provide sub-optimal results. We also notice that window shifting brings about 0.5\% improvement to the final performance. Interestingly, despite the fact that our sorting strategy does not guarantee the windows to be geometrically regular as in SST~\cite{fan2022embracing}, \model still consistently outperforms SST in all three classes. 

\myparagraph{Window Shape.}
\begin{table}[!t]
    \small\centering
    \begin{tabular}{ccccc}
        \toprule
        \multirow{2}{*}{Window Shape} & Mean L2 & Veh. L2 & Ped. L2 & Cyc. L2 \\
        & (mAPH) & (mAPH) & (mAPH) & (mAPH) \\
        \midrule
        6$\times$6 & 61.1 & 62.9 & 57.0 & 63.4 \\
        \cellcolor{gray!15}9$\times$9 & \cellcolor{gray!15}\textbf{61.7} & \cellcolor{gray!15}\textbf{63.3} & \cellcolor{gray!15}\textbf{57.9} & \cellcolor{gray!15}\textbf{63.9} \\
        13$\times$13 & 61.3 & \textbf{63.3} & \textbf{57.9} & 62.9 \\
        \bottomrule
    \end{tabular}
    \caption{\model is not sensitive to the choice of window shapes.}
    \label{tab:analysis:window_shape}
\end{table}

\model achieves robust performance under different window shapes. We choose the window shape of 9$\times$9 (2.88m$\times$2.88m, which is the size of a vehicle) in all experiments according to the results in \tab{tab:analysis:window_shape}, where we always fix the group size to be 85\% of the window shape. 

\myparagraph{Group Size.} 
\begin{table}[!t]
    \small\centering
    \begin{tabular}{ccccc}
        \toprule
        \multirow{2}{*}{Group Size} & Mean L2 & Veh. L2 & Ped. L2 & Cyc. L2 \\
        & (mAPH) & (mAPH) & (mAPH) & (mAPH) \\
        \midrule
        81$\times$50\% & 60.7 & 62.7 & 56.7 & 62.7 \\
        \cellcolor{gray!15}81$\times$85\% & \cellcolor{gray!15}\textbf{61.7} & \cellcolor{gray!15}\textbf{63.3} & \cellcolor{gray!15}\textbf{57.9} & \cellcolor{gray!15}\textbf{63.9} \\
        81$\times$125\% &  60.9 & 63.1 & 57.0 & 62.5 \\
        \bottomrule
    \end{tabular}
    \caption{Choosing a group size that is slightly smaller than the window shape (9$\times$9) provides the best accuracy on Waymo.}
    \label{tab:analysis:group}
\end{table}

We further study the choice of group sizes in \tab{tab:analysis:group}. We fix the window shape to be 9$\times$9 according to the results in \tab{tab:analysis:window_shape} and sweep the group size in 50\%, 85\% and 125\% of the window shape. The results show that setting group size to be 85\% of the window shape gives the best performance. Intuitively, if the group size is too small, \model will not be able to have a large enough receptive field (\eg, group size = 1, FWA will degenerate to MLP). When the group size is too large (say, group size = the entire point cloud), there will be a large number of outliers within each group, and \model will behave like a global PCT, which is not desired. 

\myparagraph{Input Resolution.}
\begin{table}[!t]
    \setlength{\tabcolsep}{4.5pt}
    \small\centering
    \begin{tabular}{ccccc}
        \toprule
        \multirow{2}{*}{Grid Resolution} & Mean L2 & Veh. L2 & Ped. L2 & Cyc. L2 \\
        & (mAPH) & (mAPH) & (mAPH) & (mAPH) \\
        \midrule
        0.36m & 60.7 & 63.0 & 56.7 & 62.4\\
        \cellcolor{gray!15}0.32m & \cellcolor{gray!15}\textbf{61.7} & \cellcolor{gray!15}\textbf{63.3} & \cellcolor{gray!15}\textbf{57.9} & \cellcolor{gray!15}63.9 \\
        0.28m &  \textbf{61.7} & 63.1 & 57.8 & \textbf{64.1} \\
        \bottomrule
    \end{tabular}
    \caption{Ablation on input resolution in \model: 0.32m$\times$0.32m is the best design choice that balances efficiency and accuracy.}
    \label{tab:analysis:resolution}
    \vspace{-8pt}
\end{table}

From \tab{tab:analysis:resolution}, 0.32m$\times$0.32m input resolution is the sweet spot in the latency-accuracy tradeoff for \model while further increasing the input size will only hurt the efficiency with no performance improvements.

%\myparagraph{Performance at different distances.}

\subsubsection{Breakdown for System Optimizations}
\begin{figure}[!t]
    \centering
    \includegraphics[width=\linewidth]{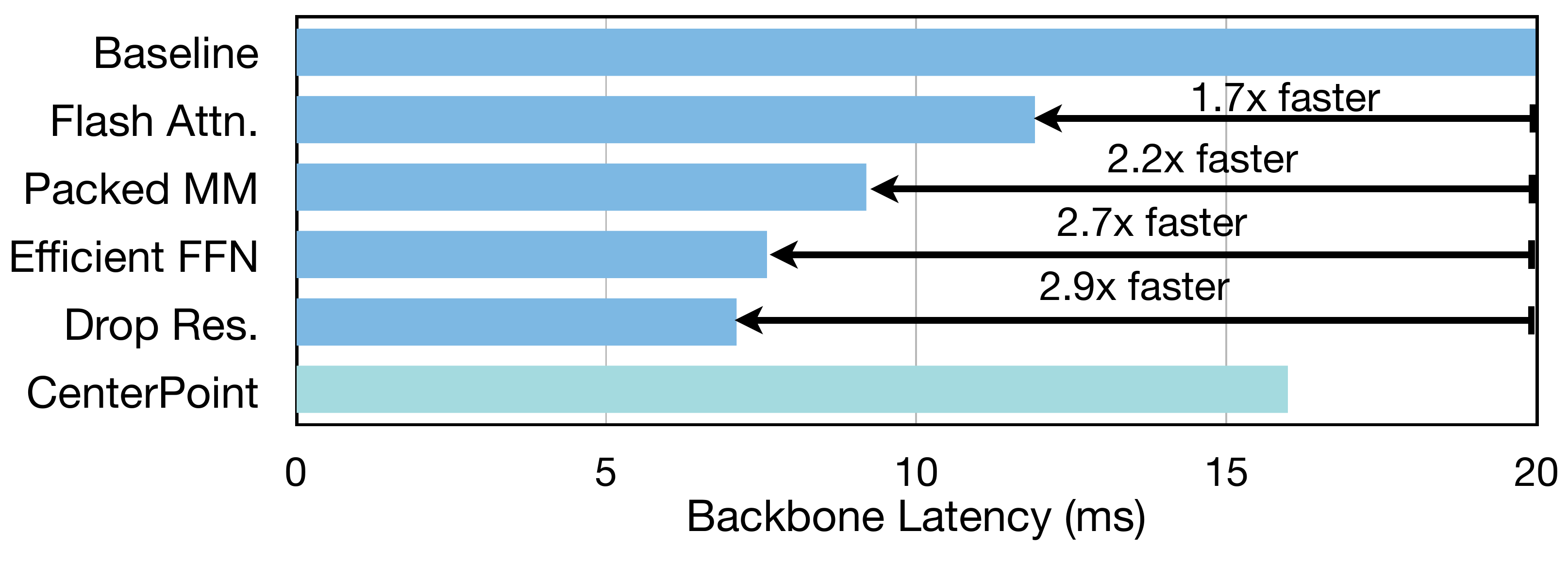}
    \caption{Improvement breakdown for system optimizations. We accelerate the backbone latency of \model by 2.9$\times$, making it \textbf{2.3$\times$} faster than CenterPoint.}
    \label{fig:analysis:system}
    \vspace{-8pt}
\end{figure}

% \paragraph{Roadmap towards a blazingly fast \model.}

In \fig{fig:analysis:system}, we analyze the effectiveness of our system optimizations proposed in \sect{sect:model:system}. An efficient MHSA implementation (FlashAttention) brings 1.7$\times$ improvement to our inference latency. Packing the computation for $\mathcal{Q}, \mathcal{K}, \mathcal{V}$ in a single linear kernel results in a 1.3$\times$ speedup. Fusing the linear and activation layers (in FFN) brings another 1.2$\times$ speedup. Finally, dropping the non-full window improves our inference latency by 1.1$\times$. To sum up, our system optimizations improve the latency of our \model by \textbf{2.9$\times$}, making its backbone \textbf{2.3$\times$ faster} than CenterPoint~\cite{yin2021center}.

\myparagraph{Discussions.}

CenterPoint is backed by SpConv~\cite{yan2018second}, which is a highly-optimized sparse convolution inference library built upon CUTLASS~\cite{nvidia2022cutlass}. Nevertheless, \model still achieves the best efficiency on NVIDIA GPUs. We partially attribute our efficiency advantage to the equally-sized groups in \model which not only gives us the best computation regularity but also eliminates the computation overhead. SpConv, on the other hand, implements 3D sparse convolution with a masked implicit GEMM algorithm, which inevitably introduces computation overhead when points within one thread block do not have exactly the same neighbor patterns. As such, \model can beat sparse convolutional models on GPUs despite their heavy system optimizations.

\section{Conclusion}

This paper introduces \model to bridge the huge efficiency gap between point cloud transformers and sparse convolutional models. It partitions the point cloud with equal-size grouping rather than equal-window grouping, trading spatial proximity for computational regularity. \model achieves state-of-the-art accuracy on Waymo Open Dataset with 4.6$\times$ speedup over previous point cloud transformers. We hope that \model can inspire future research on designing efficient and accurate point cloud transformers.
\myparagraph{Acknowledgement.}

We would like to thank Tianwei Yin, Lue Fan and Ligeng Mao for providing detailed results of CenterPoint, SST/FSD and VoTr, and Yue Wang and Yukang Chen for their helpful discussions. This work was supported by National Science Foundation, MIT-IBM Watson AI Lab, NVIDIA, Hyundai and Ford. Zhijian Liu was partially supported by the Qualcomm Innovation Fellowship.

{
\small
\bibliographystyle{ieee}
\bibliography{reference}
}

\end{document}